\documentclass{article}

\usepackage{microtype}
\usepackage{graphicx}
\usepackage{subfigure}
\usepackage{booktabs} 
\usepackage{listings}
\usepackage{xcolor}

\usepackage{hyperref}

\usepackage[accepted]{icml2025}

\usepackage{amsmath}
\usepackage{amssymb}
\usepackage{mathtools}
\usepackage{amsthm}

\usepackage[capitalize,noabbrev]{cleveref}

\theoremstyle{plain}
\newtheorem{theorem}{Theorem}[section]

\theoremstyle{definition}
\newtheorem{definition}[theorem]{Definition}

\theoremstyle{remark}

\usepackage[textsize=tiny]{todonotes}

\begin{document}

\twocolumn[
\icmltitle{STRIDE: Automating Reward Design, Deep Reinforcement Learning Training and Feedback Optimization in Humanoid Robotics Locomotion}



\icmlsetsymbol{equal}{*}

\begin{icmlauthorlist}
\icmlauthor{Zhenwei Wu}{equal,comp,sch}
\icmlauthor{Jinxiong Lu}{equal,comp}
\icmlauthor{Yuxiao Chen}{comp}
\icmlauthor{Yunxin Liu}{sch2}
\icmlauthor{Yueting Zhuang}{sch3}
\icmlauthor{Luhui Hu}{comp}
\end{icmlauthorlist}

\icmlaffiliation{comp}{ZhiCheng AI, Hangzhou, China}
\icmlaffiliation{sch}{School of Automation Science and Engineering, South China University of Technology, Guangzhou, China}
\icmlaffiliation{sch2}{Institute for AI Industry Research, Tsinghua University, Beijing, China}
\icmlaffiliation{sch3}{College of Computer Science and Technology, Zhejiang University, Hangzhou, China}



\vskip 0.3in
]



\printAffiliationsAndNotice{\icmlEqualContribution}  

\begin{abstract}
	Humanoid robotics presents significant challenges in artificial intelligence, requiring precise coordination and control of high-degree-of-freedom systems. Designing effective reward functions for deep reinforcement learning (DRL) in this domain remains a critical bottleneck, demanding extensive manual effort, domain expertise, and iterative refinement. To overcome these challenges, we introduce STRIDE, a novel framework built on agentic engineering to automate reward design, DRL training, and feedback optimization for humanoid robot locomotion tasks. By combining the structured principles of agentic engineering with large language models (LLMs) for code-writing, zero-shot generation, and in-context optimization, STRIDE generates, evaluates, and iteratively refines reward functions without relying on task-specific prompts or templates. Across diverse environments featuring humanoid robot morphologies, STRIDE outperforms the state-of-the-art reward design framework EUREKA, achieving an average improvement of round $250\%$ in efficiency and task performance. Using STRIDE-generated rewards, simulated humanoid robots achieve sprint-level locomotion across complex terrains, highlighting its ability to advance DRL workflows and humanoid robotics research.
\end{abstract}

\begin{figure*}[t]
	\centering
	\includegraphics[width=0.99\linewidth]{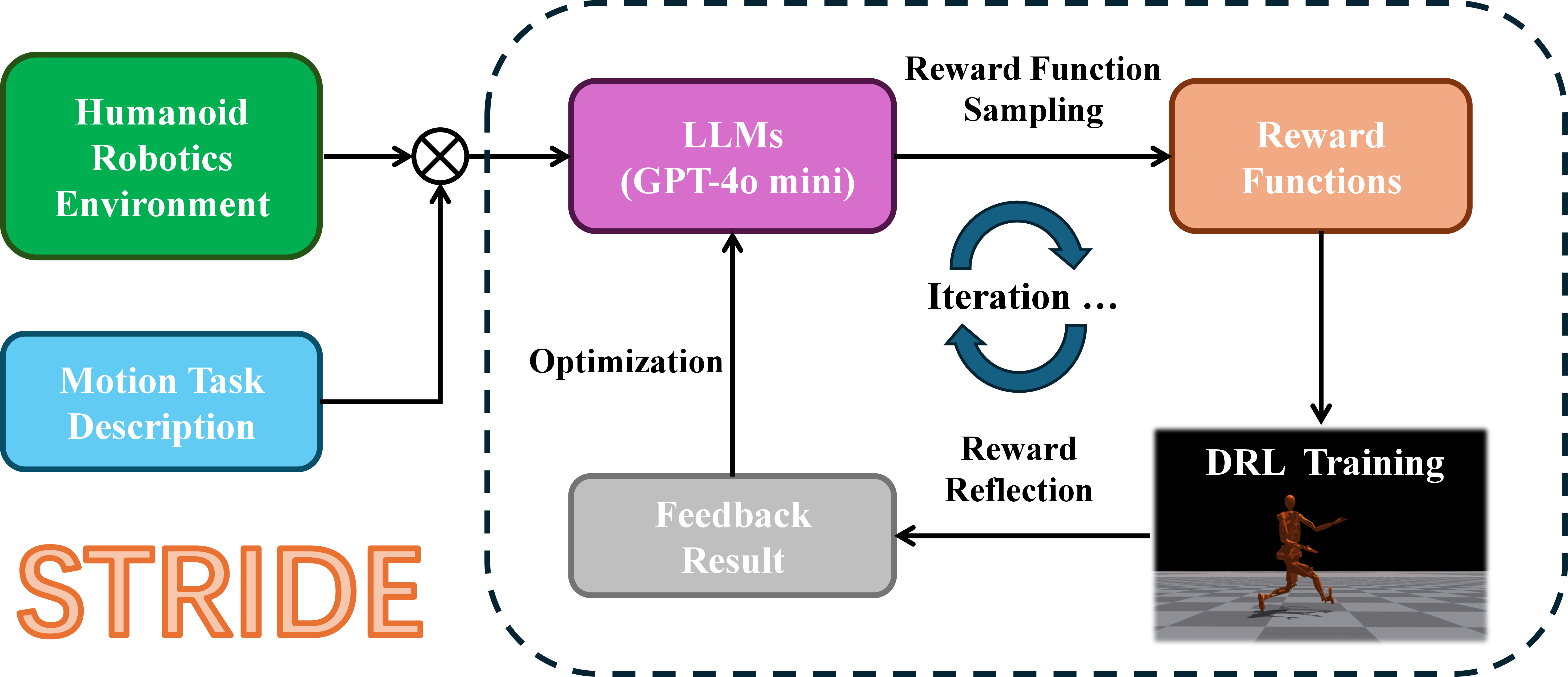}
	\caption{STRIDE pipeline: The framework integrates environment code, task descriptions, LLMs, and reinforcement learning to automate reward generation and optimization for humanoid robot locomotion tasks.}
	\label{Framework}
\end{figure*}

\section{Introduction}
\label{introduction}

Humanoid robotics represents a pinnacle of artificial intelligence and control systems, requiring solutions for high-dimensional, dynamic tasks such as walking, running, and maintaining balance. Among these challenges, achieving human-like athletic locomotion—such as sprinting—stands as a benchmark for progress in robot control and coordination. This goal demands not only precise motor control and stability but also dynamic adaptability to varying terrains and external disturbances. Traditional reinforcement learning (RL) methods have made significant strides in training agents for humanoid tasks; however, these approaches heavily rely on human-engineered reward functions. Designing such rewards requires substantial domain expertise and meticulous tuning, limiting their scalability and adaptability across diverse environments and complex tasks \cite{cao2024survey}. Moreover, human-designed rewards often reflect implicit biases and may fail to uncover optimal solutions, particularly in scenarios involving high degrees of freedom. These limitations underscore the need for automated methods capable of designing robust reward functions for complex tasks \cite{zeng2024learning}.

Early research has laid the groundwork for automated reward function learning. \cite{albilani2022dynamic}, \cite{tang2024humanmimic} and \cite{zare2024survey} explored reward function generation through imitation learning and evolutionary algorithms, while \cite{li2021reinforcement}, \cite{weng2021natural} and \cite{radosavovic2024real} demonstrated the potential of RL in teaching humanoid robots to walk or mimic predefined trajectories. However, these approaches either lack generality or require predefined human knowledge for reward construction. Applying LLMs to directly generate and iteratively improve reward functions for complex tasks, such as humanoid sprinting, remains underexplored.

Recent advancements in LLMs, such as GPT-4, have demonstrated remarkable capabilities in high-level reasoning, zero-shot generation, and in-context optimization \cite{naveed2023comprehensive,qin2023cross,chen2024large}. These features position LLMs as transformative tools for addressing the challenges of RL reward engineering, particularly by automating the design, evaluation, and iterative refinement of reward functions \cite{jiang2024survey,kepel2024autonomous}. Unlike traditional approaches, which rely on fixed templates or manual design, LLMs can dynamically interpret task descriptions and environment constraints to produce executable reward functions tailored to specific objectives \cite{kim2023language,han2024autoreward,ma2024eureka}. This flexibility opens the door to solving high-dimensional control problems, including humanoid sprinting, in ways previously unattainable with conventional methods.

In this work, we introduce STRIDE, an AI-driven framework designed to automate reward function generation, DRL training, and iterative feedback optimization for humanoid robot locomotion. STRIDE leverages the advanced code generation, in-context optimization, and reasoning capabilities of LLMs to evolve reward functions without requiring task-specific prompts or predefined templates. By incorporating feedback from DRL training outcomes, STRIDE continuously refines reward designs, outperforming state-of-the-art frameworks like Eureka, with round $250\%$ improvement in overall performance across various robot morphologies and locomotion tasks. STRIDE consistently delivers superior results compared to other methods, significantly improving task performance in diverse moving environments.

Furthermore, STRIDE introduces a gradient-free integration of human feedback, enabling the generation of safer, higher-quality rewards without additional retraining. This flexibility significantly enhances its applicability to real-world scenarios where safety and reward quality are paramount. Finally, applying STRIDE-generated rewards in a curriculum learning setting enables simulated humanoid robots to sprint with athletic precision, mimicking the dynamics of professional sprinters. By bridging the gap between LLM capabilities and humanoid robot locomotion, STRIDE sets a new standard for automation in robot control and reward engineering.

\section{Related Work}
\label{Related_Work}

\textbf{Reward Design in Reinforcement Learning.}
Reward design is a fundamental aspect of RL, directly influencing an agent’s ability to learn effective policies. Traditional approaches rely heavily on hand-crafted rewards tailored to specific tasks \cite{mo2022managing}. While these methods provide clarity and control, they often lack adaptability across diverse environments and require significant domain expertise. To address these limitations, automated reward generation has emerged as a key area of research. For example, \cite{sarukkai2024automated} proposes an LLMs-driven framework that uses progress functions and intrinsic rewards to efficiently generate rewards for reinforcement learning tasks. \cite{katara2024gen2sim} introduces Gen2Sim, which automates 3D asset creation, tasks, and rewards to scale robot learning in simulation. However, these approaches primarily focus on low-dimensional tasks, and their applicability to high-dimensional systems, such as humanoid robot locomotion, remains constrained.

\textbf{Reinforcement Learning for Humanoid Robotics.}  
Humanoid robots pose significant challenges for reinforcement learning (RL) due to their high degrees of freedom and the necessity for dynamic stability. Early research on humanoid robot locomotion utilized model-based methods, such as the inverted pendulum approach \cite{nandula2021neurodynamic}, which proved insufficient for handling complex movements like sprinting. More recent advancements have employed RL to develop motor skills in humanoid robots. For instance, \cite{tang2024humanmimic} introduced a Wasserstein adversarial imitation learning system, enabling humanoid robots to replicate human locomotion and execute seamless transitions. Similarly, \cite{pei2024gait} proposed a DDPG-based control framework with optimized reward functions to achieve stable, high-speed, and humanlike bipedal walking. Additionally, \cite{figueroa2024reinforcement} integrated intrinsically stable model predictive control and whole-body admittance control into an RL framework to enhance bipedal walking stability under dynamic conditions. While these approaches mark significant progress, achieving sprint-level performance in humanoid robotics still demands more advanced reward engineering and sophisticated learning algorithms.

\textbf{Large Language Models in RL and Control Tasks.}
LLMs, such as GPT-4, have demonstrated exceptional capabilities in zero-shot reasoning, code generation, and contextual optimization. Recent studies have begun exploring their potential in RL reward design. For instance, \cite{li2024auto} introduces Auto MC-Reward, an LLM-driven system that designs and refines dense reward functions to improve reinforcement learning efficiency. \cite{beak2024chatpcg} proposes ChatPCG, an LLM-driven framework that automates reward design and content generation for multiplayer game AI development. However, most of these efforts focus on high-level planning tasks, with limited exploration in low-level control for high-dimensional systems. Integrating LLMs into RL pipelines for humanoid robotics remains a largely untapped area of research.

\textbf{Feedback-Driven Optimization in RL.}
Iterative feedback has proven to be a powerful tool in RL for refining policies and reward functions. Frameworks like Reinforcement Learning from Human Feedback (RLHF) have demonstrated success in incorporating human input to improve model performance. However, RLHF methods typically require manual feedback, making them resource-intensive and less scalable. Recent advancements, such as the adapting feedback-driven DRL algorithm \cite{pattnaik2021multitask}, the online data-driven model-based inverse RL technique \cite{self2022model} and the dynamic inverse RL method \cite{tan2024dynamic} have shown promise in automating this process, but they remain insufficiently explored in the context of humanoid robotics locomotion.

\textbf{Positioning of Our Work.}
Our work builds on these foundational efforts by introducing STRIDE, a framework that leverages LLMs for fully automated reward generation, training, and feedback-driven optimization. Unlike traditional methods that rely on task-specific templates or human intervention, STRIDE dynamically generates rewards tailored to complex tasks like humanoid sprinting. By integrating LLMs' generative capabilities with reinforcement learning, STRIDE bridges the gap between high-level planning and low-level control, outperforming human-engineered rewards and state-of-the-art frameworks such as Eureka. Furthermore, its gradient-free feedback mechanism ensures safer and higher-quality rewards, setting a new benchmark for humanoid robotics and RL reward engineering.

\section{Problem Setting and Definitions}

The goal of reward design is to generate a shaped reward function for a ground-truth reward function that may be difficult to optimize directly (e.g., sparse rewards). This ground-truth reward function can only be accessed through queries by the designer. To formalize the problem, we adapt the reward design definition from \cite{singh2009rewards} to our program synthesis setting, which we term \textit{reward generation}.

\begin{definition}[Reward Design Problem (RDP), adapted from \cite{singh2009rewards}]
	A reward design problem is defined as a tuple $P = \langle M, \mathcal{R}, \pi_M, F \rangle$, 
	
	where $M = (S, A, T)$ is the \textit{world model}, consisting of: $S$ represents the state space, $A$ represents the action space, $T$ represents the transition function. $\mathcal{R}$ is the space of reward functions. $A_M(\cdot): \mathcal{R} \rightarrow \Pi$ is a learning algorithm that outputs a policy $\pi: S \rightarrow \Delta(A)$, which optimizes a given reward function $R \in \mathcal{R}$ in the resulting Markov Decision Process (MDP) $(M, R)$. $F: \Pi \rightarrow \mathbb{R}$ is the fitness function, providing a scalar evaluation of a policy, accessible only through policy queries (i.e., evaluating the policy using the fitness function).
	
	The objective of an RDP is to output a reward function $R \in \mathcal{R}$ such that the policy $\pi := A_M(R)$ achieves the highest fitness score $F(\pi)$.
\end{definition}

\paragraph{Reward Generation Problem.} 
In our setting, every component of an RDP is specified via code. Given a task string $l$ that describes the objective, the goal of the reward generation problem is to output a reward function code $R$ such that $F(A_M(R))$ is maximized.

\section{Method}
\subsection{Experimental Setup}

\begin{figure*}[t]
	\centering
	\includegraphics[width=1\linewidth]{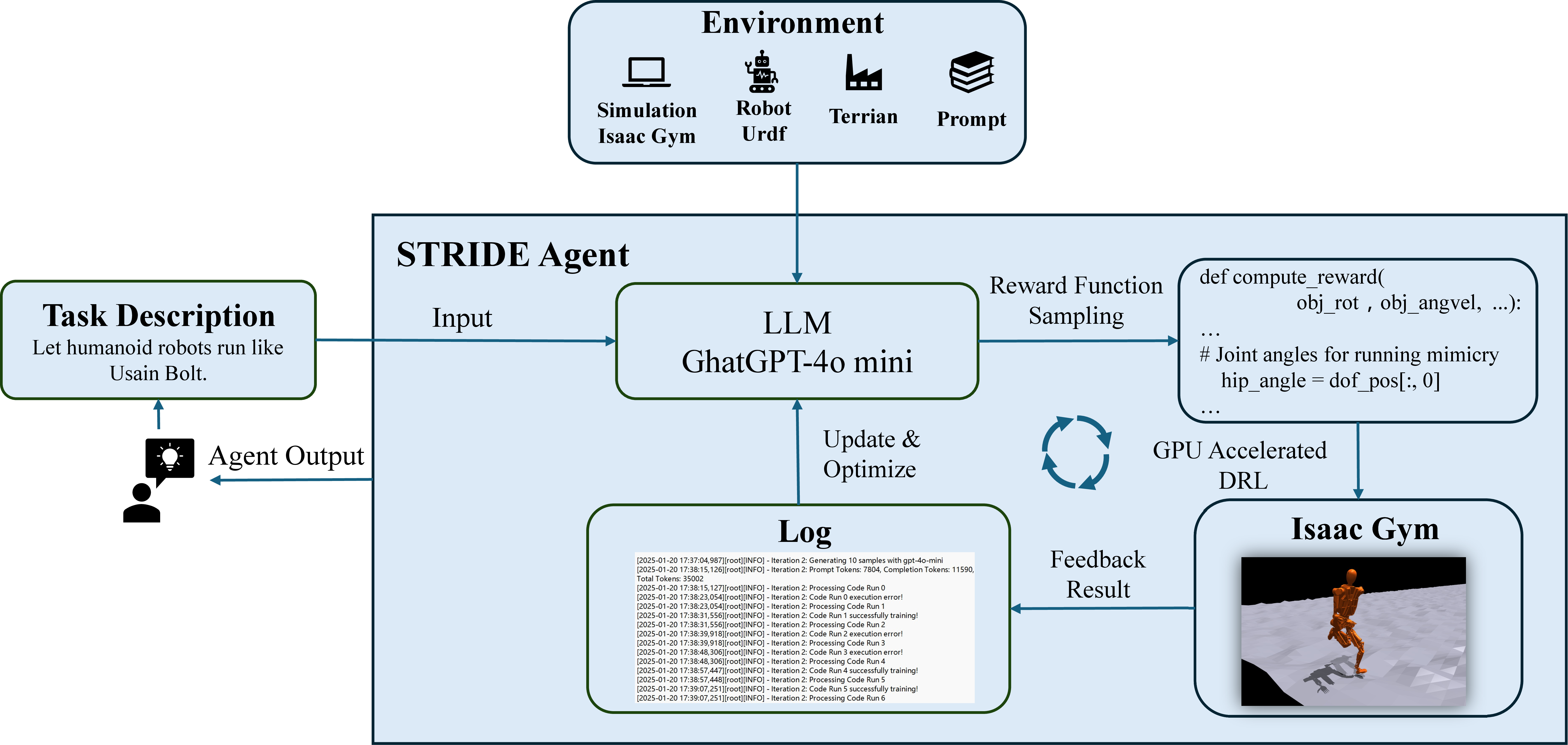}
	\caption{STRIDE Agent Framework: Integrating Environment Code, Task Descriptions, LLMs, and DRL for Automated Reward Generation and Optimization in Humanoid Robot Locomotion.}
	\label{agent}
\end{figure*}

This section presents our proposed agent framework (Figure \ref{agent}), STRIDE, which automates reward design, iterative optimization, and reinforcement learning training for humanoid robot locomotion tasks. STRIDE leverages the code generation capabilities of LLMs to create and refine reward functions dynamically, without relying on task-specific templates or manual intervention. The process consists of three main components: (1) Environment as Context, (2) Feedback-Driven Reward Optimization, and (3) Reinforcement Learning Training. Figure~\ref{Framework} provides an overview of the pipeline.

\subsection{Overview}
STRIDE begins by providing the raw environment code and task descriptions as context to an LLM, which generates reward functions in Python. These reward functions are evaluated by training agents in a simulated environment, with their performance scored using a predefined fitness function. STRIDE iteratively refines the reward functions by incorporating textual feedback generated from training results. This closed-loop process continues until an optimal reward function is obtained. The final reward function is then used to train humanoid robots, such as sprinting agents, to achieve task-specific goals with improved performance and stability.

\subsection{Environment as Context}
Reward design begins with providing the raw environment code and task description as input context to the LLM. STRIDE dynamically adapts to the environment by parsing the variables, constraints, and structure specified in the code. This enables the framework to generate reward functions that are semantically aligned with the specific characteristics of the task and environment, eliminating the need for task-specific templates.

\paragraph{Humanoid Robot Description.}
The humanoid robot used in our experiments is custom-designed in SolidWorks and exported as a URDF file. This robot features a total of 16 degrees of freedom (DoF), enabling complex movements required for tasks such as sprinting, balancing, and navigating uneven terrains. The robot's joints include:
\begin{itemize}
	\item \textbf{Lower Body:} Hip roll/pitch/yaw (3 DoF per leg), knee pitch (1 DoF per leg), and ankle pitch/roll (2 DoF per leg).
	\item \textbf{Upper Body:} Shoulder flexion/extension (1 DoF per arm), elbow flexion/extension (1 DoF per arm).
\end{itemize}
The robot's morphology and joint configuration allow it to mimic human-like movements while maintaining stability during dynamic tasks. These details are explicitly encoded in the URDF file and parsed by STRIDE to dynamically extract relevant state variables, such as joint angles, velocities, and torso alignment, for use in reward generation.

\paragraph{Why Humanoid Details Matter.}
Integrating the humanoid robot’s detailed URDF configuration into STRIDE ensures that the reward functions are tailored to the robot’s specific capabilities. For instance, STRIDE leverages the joint hierarchy and kinematic constraints defined in the URDF to design rewards that promote efficient locomotion and balance. This approach enhances the adaptability of the framework across diverse humanoid designs, ensuring that STRIDE-generated rewards are optimized for the robot’s unique physical structure.

By combining the detailed humanoid description with the raw environment code, STRIDE fully automates the reward design process, making it scalable to other robot morphologies and tasks. This capability demonstrates the versatility of STRIDE in leveraging agentic engineering to advance humanoid robot locomotion and reward engineering.

\paragraph{Why Environment Code?}
Using the environment code as context has two major advantages:
\begin{enumerate}
	\item \textbf{Compatibility with LLM Training.} LLMs are pretrained on code in existing programming languages. By directly feeding the raw environment code, STRIDE enables the LLM to generate reward functions in the same syntax and style, improving the quality of the generated code.
	\item \textbf{Dynamic Adaptability.} Environment code explicitly defines the variables and observations available for reward function design. This ensures that the LLM-generated rewards are grounded in the semantics of the environment, avoiding reliance on hard-coded templates.
\end{enumerate}

For example, in a humanoid sprinting task, the LLM automatically identifies key variables such as joint velocities, torso positions, and step frequencies from the environment's observation space, leveraging them to construct effective reward functions.

\subsection{Feedback-Driven Reward Optimization}
While zero-shot reward generation from the LLM can yield plausible reward functions, these initial designs are not always optimal or executable. To address this, STRIDE introduces an iterative feedback-driven optimization loop.

\paragraph{Reward Evaluation.} After generating reward functions, STRIDE evaluates them by integrating them into a reinforcement learning pipeline. The fitness of each reward function is computed based on the agent's performance, such as sprint speed, stability, and energy efficiency.

\paragraph{Reward Mutation and Refinement.} Using the evaluation results, STRIDE refines reward functions by providing detailed textual feedback to the LLM. This feedback includes:
\begin{itemize}
	\item Adjustments to hyperparameters, such as scaling coefficients for reward components.
	\item Modifications to reward terms to enhance credit assignment for specific behaviors.
	\item Suggestions for introducing new reward components based on training deficiencies.
\end{itemize}

\subsection{DRL Training}
Once an optimized reward function is obtained, STRIDE uses it to train policies for humanoid robot locomotion tasks. The training process occurs in a high-performance simulation environment, such as Isaac Gym, which supports parallelized training of multiple agents. The trained policies are evaluated on their ability to achieve task-specific objectives, such as high-speed sprinting or maintaining balance during dynamic movements.

\paragraph{Scalability and Efficiency.} By automating reward design and optimization, STRIDE significantly reduces the manual effort required in traditional RL pipelines. Its compatibility with diverse environments ensures scalability to a wide range of humanoid robotics tasks.

\begin{algorithm}[t]
	\caption{STRIDE Framework}
	\label{alg:stride}
	\begin{algorithmic}[1]
		\REQUIRE Humanoid robotics environment code $M$, motion task description $l$, coding LLM $LLM$, reward function $F$, initial prompt $prompt$
		\STATE Initialize reward function $R_\text{best}$ and fitness $s_\text{best} = 0$
		\FOR{each iteration $n = 1, 2, \ldots, N$}
		\STATE Generate $K$ reward functions: $R_1, \ldots, R_K \sim LLM(l, M, prompt)$
		\STATE Evaluate rewards: $s_1, \ldots, s_K = F(R_1), \ldots, F(R_K)$
		\STATE Update best reward: $(R_\text{best}, s_\text{best}) = \text{argmax}_{i}(R_i, s_i)$
		\STATE Refine prompt using feedback: $prompt \leftarrow \text{Reflection}(R_\text{best}, s_\text{best})$
		\STATE \textbf{Return} Optimized reward function $R_\text{best}$
		\ENDFOR
	\end{algorithmic}
\end{algorithm}

\subsection{Illustration of the Pipeline}
Figure~\ref{Framework} illustrates the overall STRIDE framework. The environment code and task description are first processed by the LLM to generate an initial reward function. This reward function is integrated into the RL training pipeline, and feedback is used to iteratively refine the reward design. The process continues until the optimal reward function is found, which is then used to train policies for humanoid robot locomotion tasks.

%
%
%

\section{Experiments}
We comprehensively evaluate STRIDE across a diverse suite of humanoid robot locomotion tasks and terrains, assessing its ability to automate reward design, optimize DRL training, and iteratively refine policies based on feedback. STRIDE leverages GPT-4o mini as its underlying LLM, augmented by agentic engineering principles to seamlessly integrate environment constraints, motion task descriptions, and iterative feedback loops. The results highlight STRIDE's superior performance over existing frameworks EUREKA \cite{ma2024eureka}. In all our experiments, STRIDE conducts 10 independent runs per environment, and for each run, searches for 3 iterations with $K = 10$ samples per iteration.

\subsection{Experimental Setup}
\begin{figure}[t]
	\centering
	\includegraphics[width=1\linewidth]{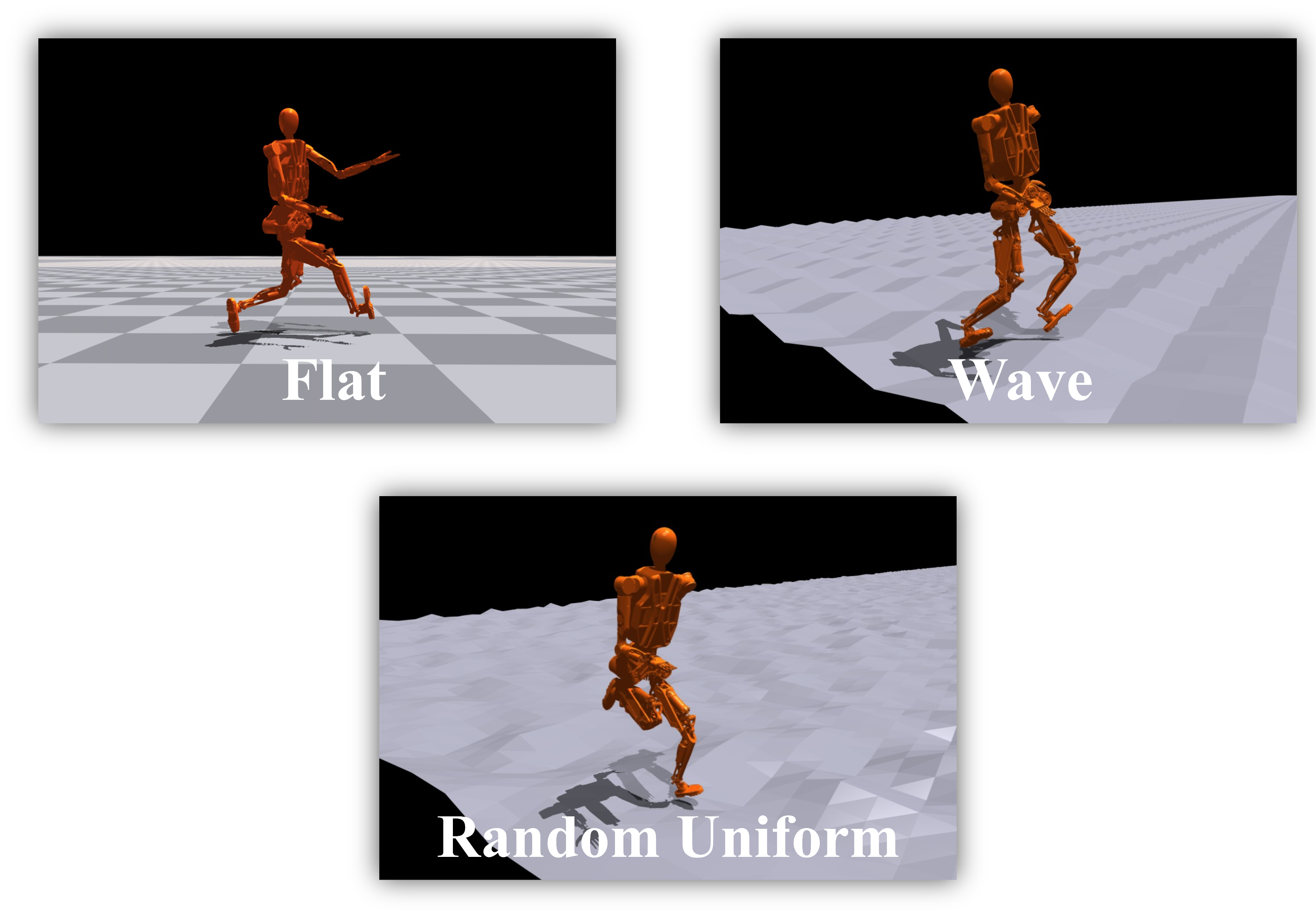}
	\caption{Humanoid robots test on different terrains.}
	\label{terrains}
\end{figure}
\textbf{Environments:}
Our experiments are conducted in the Isaac Gym simulator \cite{makoviychuk2021isaac}, featuring a suite of humanoid robot locomotion tasks designed to test performance under varying levels of complexity. As shown in Figure \ref{terrains}, we evaluate STRIDE on:

1. \textbf{Flat Terrain}: Testing the robot’s ability to achieve maximum sprint speed on uniform ground.

2. \textbf{Wave Terrain}: Simulating uneven wave-like surfaces to evaluate adaptability.

3. \textbf{Random Uniform Terrain}: Introducing random height variations to assess robustness in dynamic environments.

\textbf{Baselines:} EUREKA, a state-of-the-art reward design framework using GPT-4 for generating reward functions.

\textbf{Evaluation Metrics: }
For each task, we measure the Max Success Score that evaluates the maximum task completion rate across five independent DRL training runs. Moreover, we also measured the success rate of generating code. 

\subsection{Results}
\begin{figure*}[htp!]
	\centering
	\includegraphics[width=0.95\linewidth]{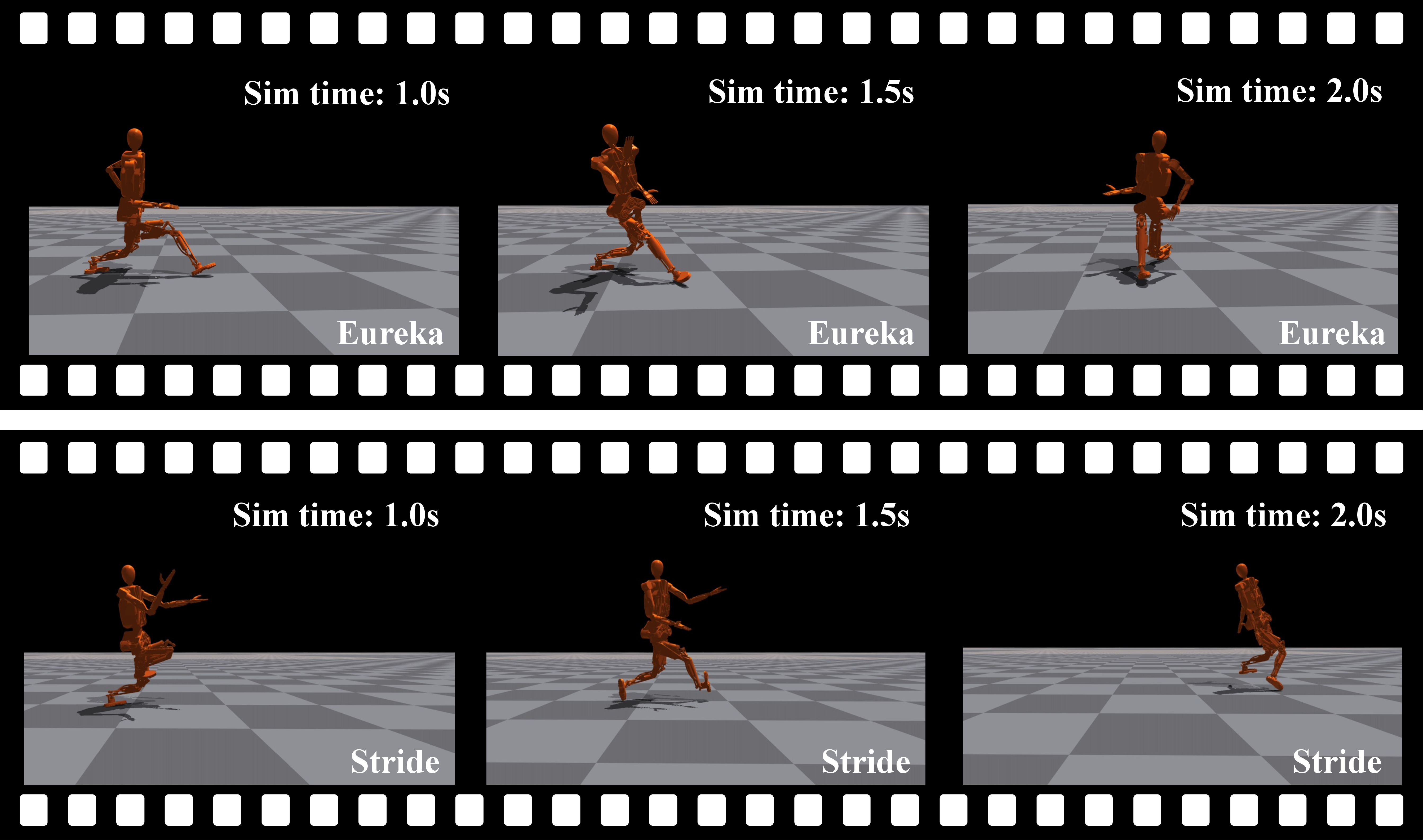}
	\caption{Comparison of STRIDE and EUREKA on the flat terrain.}
	\label{contrast}
\end{figure*}

\begin{figure}[t]
	\centering
	\includegraphics[width=1\linewidth]{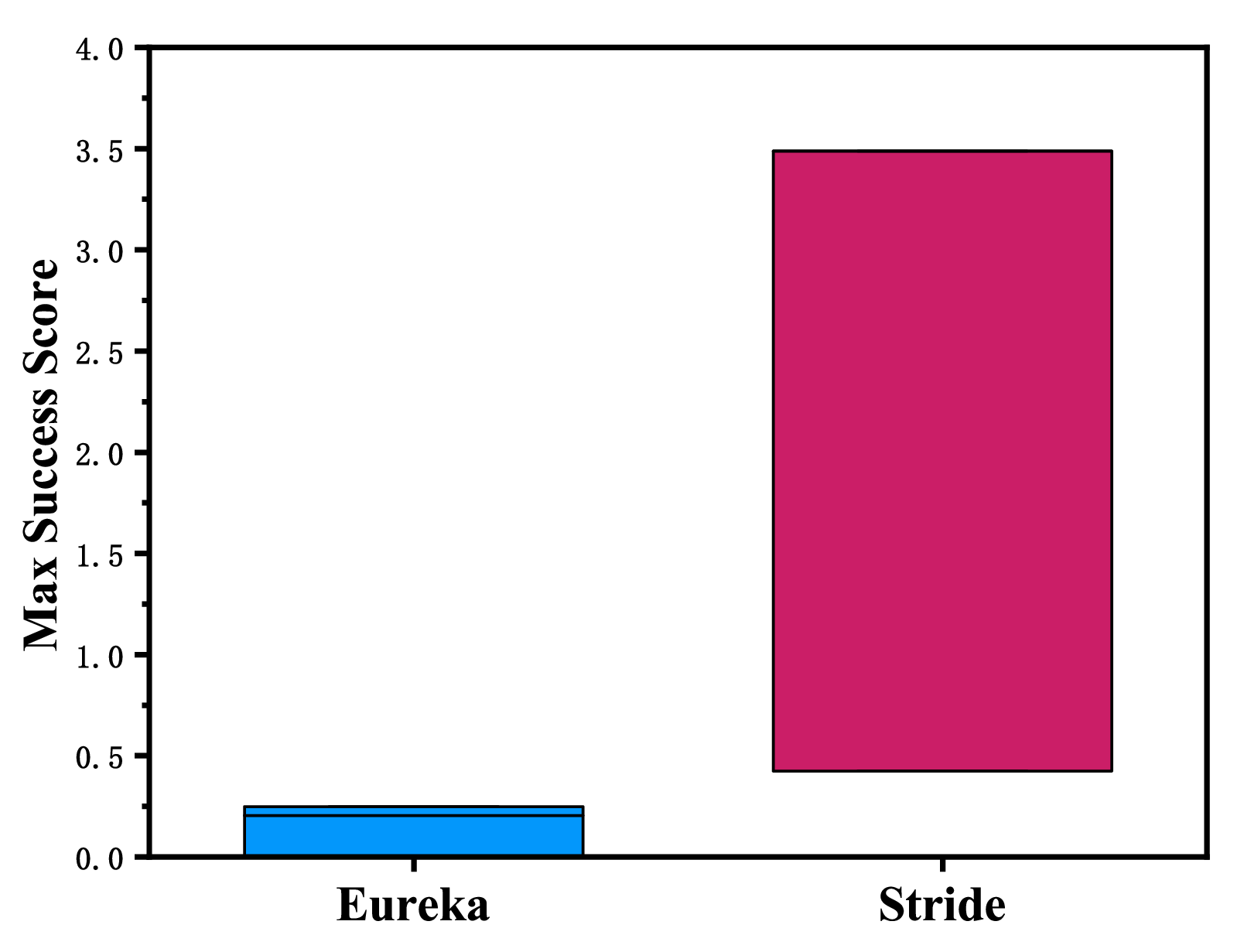}
	\caption{Comparison of Stride and Eureka on the flat terrains.}
	\label{flat}
\end{figure}

\begin{figure}[t]
	\centering
	\includegraphics[width=1\linewidth]{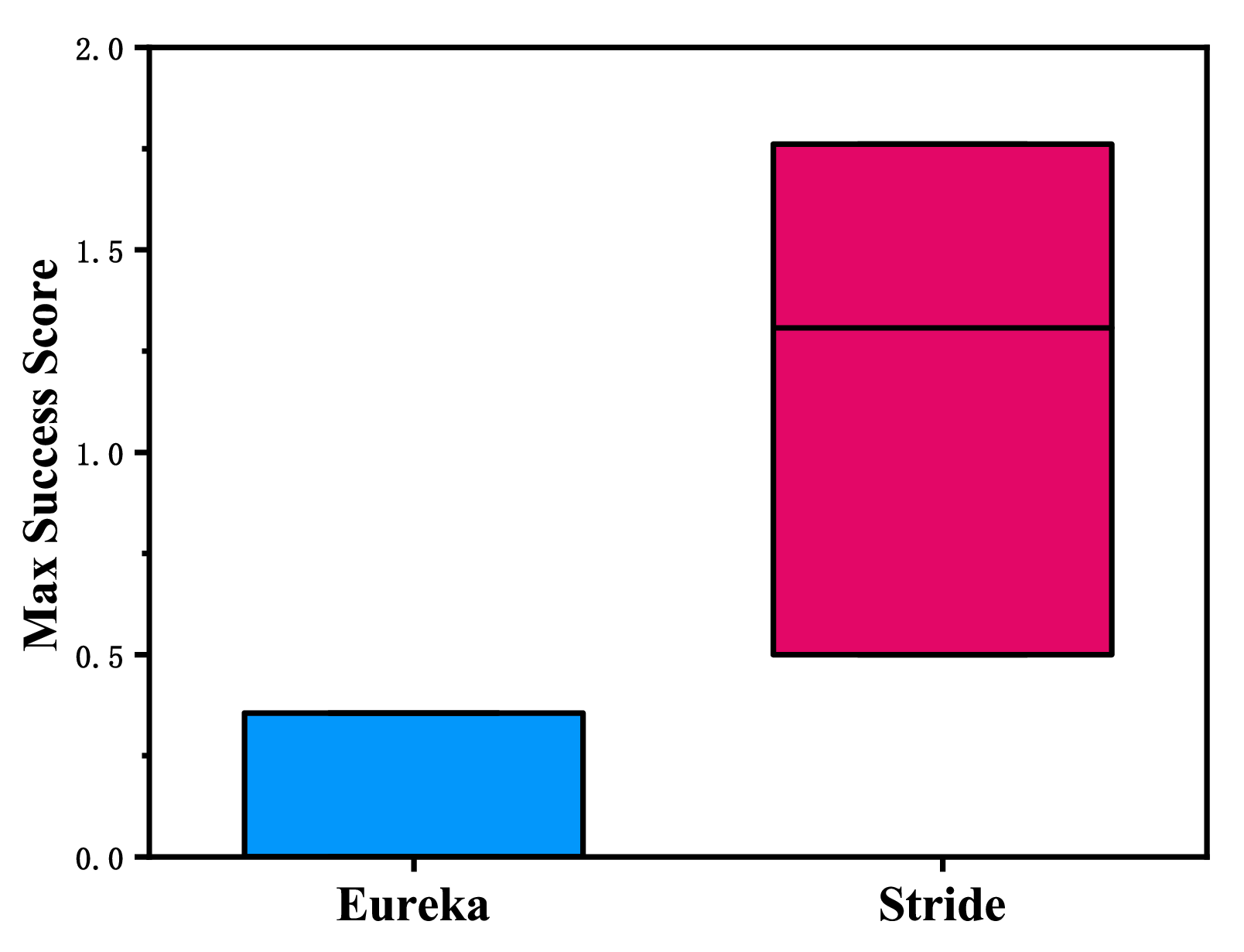}
	\caption{Comparison of Stride and Eureka on wave terrain.}
	\label{wave}
\end{figure}

\begin{figure}[t]
	\centering
	\includegraphics[width=1\linewidth]{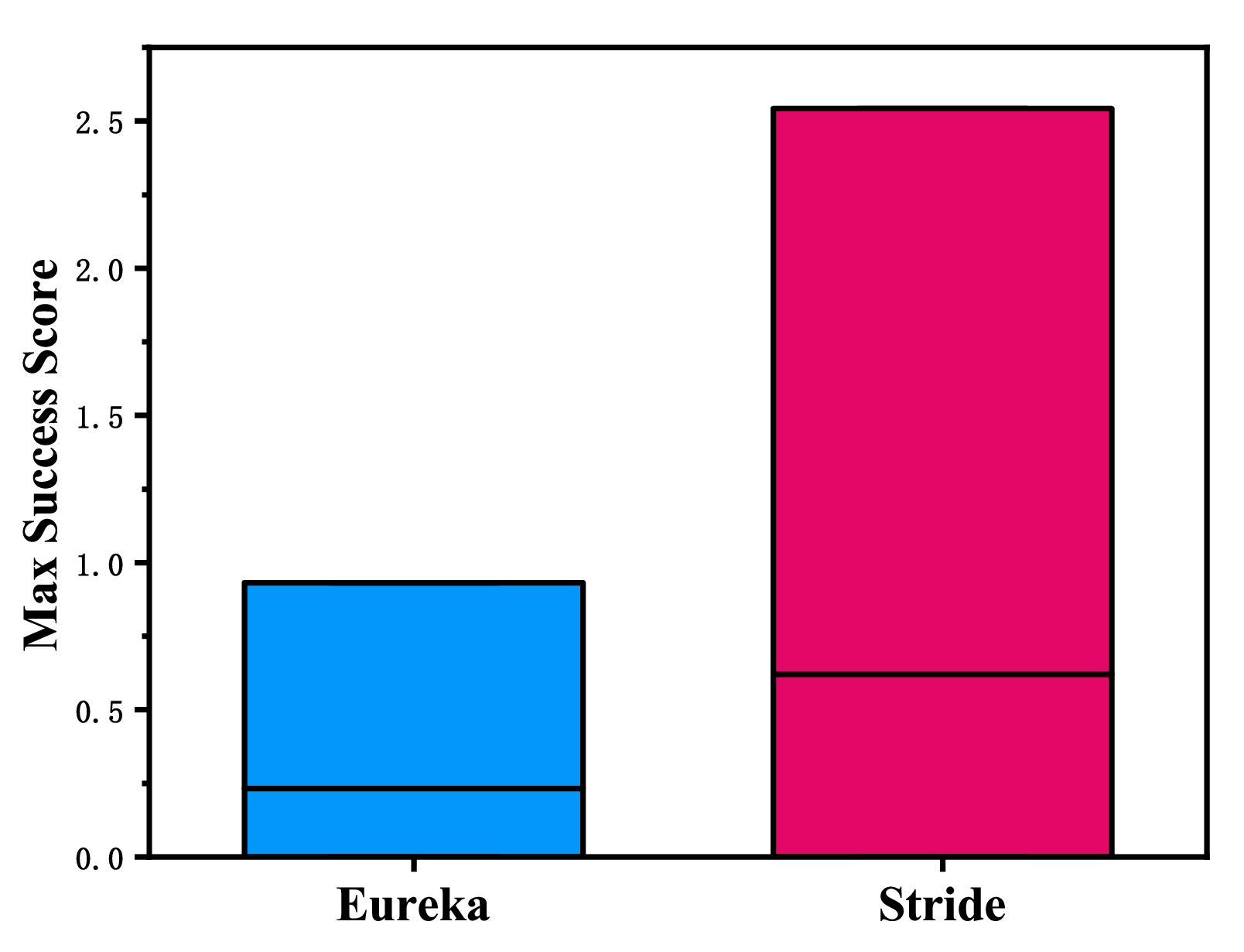}
	\caption{Comparison of Stride and Eureka on random uniform terrain.}
	\label{random}
\end{figure}

\textbf{STRIDE outperforms EUREKA.}
Figure \ref{contrast} demonstrates that Stride significantly outperforms Eureka in humanoid robot sprinting. While Eureka produces awkward and inefficient running with poor balance and coordination, Stride generates a more natural, fluid, and stable sprinting motion. At each time step (1s, 1.5s, and 2s), the robot using Stride shows better joint coordination and posture, indicating that Stride’s reward function is better optimized for humanoid locomotion. This comparison highlights Stride's superior ability to generate effective rewards for complex tasks like humanoid sprinting.The STRIDE-generated rewards adaptively balance speed and stability by incorporating joint velocity, torso alignment, and energy efficiency terms. The resulting policy enables the humanoid robot to achieve consistent sprinting behavior while maintaining balance, outperforming EUREKA rewards by a wide margin. Figure \ref{flat}-\ref{random} present the results across three terrains. STRIDE consistently exceeds the performance of EUREKA baselines:
\begin{itemize}
	\item On flat terrain, STRIDE achieves near-optimal sprinting speed, outperforming EUREKA (Figure~\ref{flat}).
	
	\item On wave terrain, STRIDE demonstrates superior adaptability, achieving three times the Max Success Score of EUREKA (Figure~\ref{wave}).
	
	\item On random uniform terrain, STRIDE achieves unprecedented robustness, with the Max Success Score exceeding round $250\%$ of EUREKA’s best performance (Figure~\ref{random}). 
\end{itemize}

Table \ref{success_rate} presents the success rates of Eureka, Stride, and Stride with Human Initialization across three iterations. Eureka starts with a low success rate of 0.1, improves slightly to 0.4 in the second iteration, but drops to 0.0 in the third iteration, demonstrating its inability to improve consistently. Stride shows gradual improvement from 0.2 in the first iteration to 0.5 in the second, but its progress stagnates at 0.3 in the third iteration. In contrast, Stride with Human Initialization maintains a high and consistent success rate of 0.7 across all iterations, highlighting the advantage of human-initialized reward functions in achieving stable and superior performance.

\begin{table}[t]
	\caption{Success rates for Eureka, Stride and Stride (human init.) on different iterations.}
	\label{success_rate}
	\vskip 0.15in
	\begin{center}
		\begin{small}
			\begin{sc}
				\begin{tabular}{lcccr}
					\toprule
					Iteration & Eureka & Stride & \textbf{Stride (human init.)} \\
					\midrule
					1    	  & 0.1		& 0.2   & \textbf{0.7} \\
					2 	 	  & 0.4		& 0.5   & \textbf{0.7} \\
					3    	  & 0.0		& 0.3   & \textbf{0.7} \\
					\bottomrule
				\end{tabular}
			\end{sc}
		\end{small}
	\end{center}
	\vskip -0.1in
\end{table}

\subsection{STRIDE with Human Init.}
\begin{figure}[t]
	\centering
	\includegraphics[width=1\linewidth]{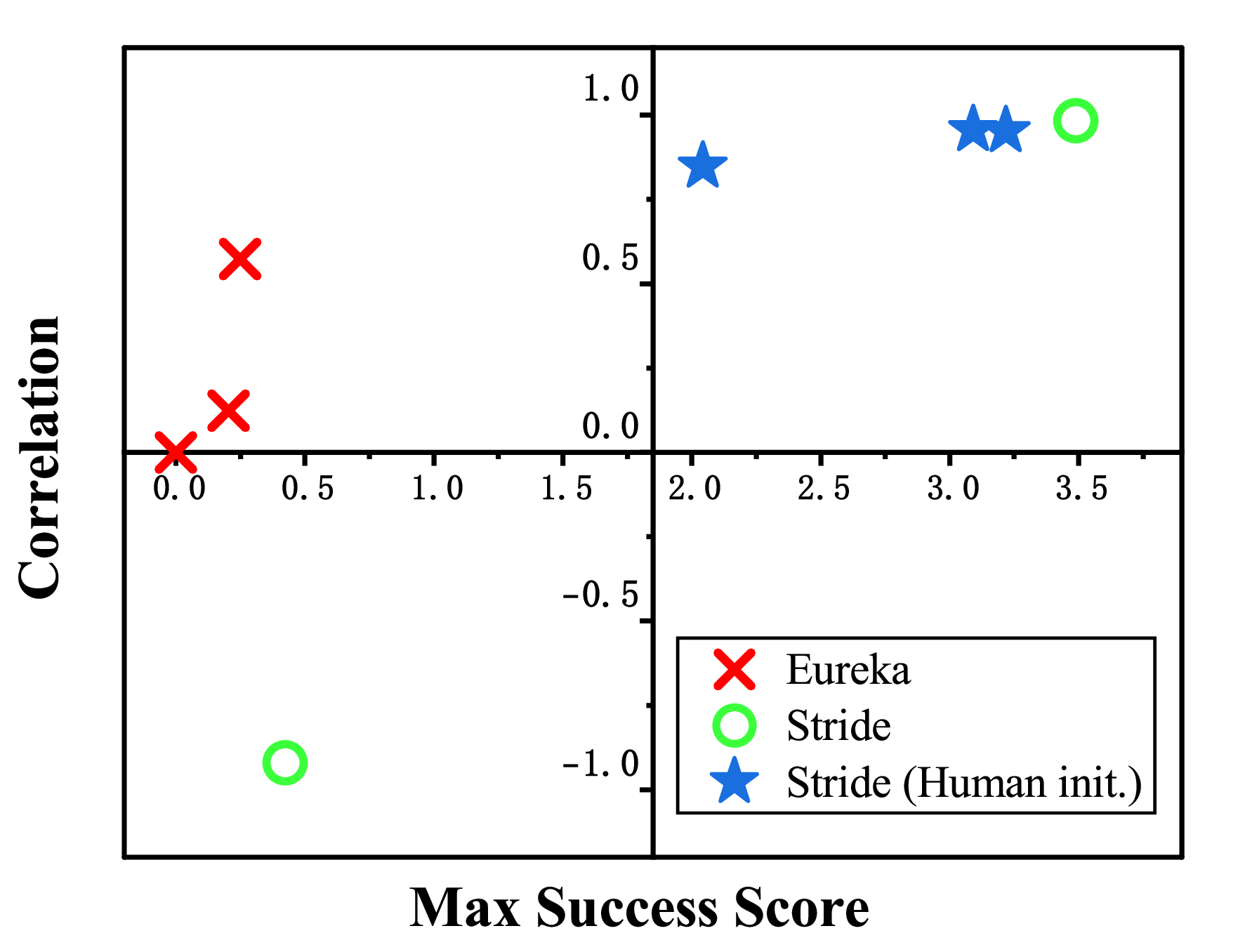}
	\caption{Correlation analysis of Eureka, Stride and Stride (Human init.).}
	\label{correlation}
\end{figure}

Finally, we investigate STRIDE's ability to integrate human init. to refine its rewards. Using a gradient-free approach, we allow human users to describe desired improvements (e.g., prioritizing stability over speed). The results, visualized in Figure~\ref{correlation}, show that Stride (particularly with human initialization) outperforms Eureka both in terms of success score and alignment with human-designed rewards. This suggests that Stride can generate more effective and innovative reward functions while still being able to integrate human expertise for improved outcomes. The analysis also emphasizes that Stride's ability to generate novel and useful reward functions might not always correlate directly with human expectations, but it still drives superior performance in humanoid robotics tasks.

This experimental analysis demonstrates STRIDE’s superior performance, adaptability, and capability to innovate beyond human intuition, solidifying its role as a groundbreaking framework for automated reward design in humanoid robot locomotion tasks.


\section{Conclusion}
We have introduced \textbf{STRIDE}, a novel framework that leverages agentic engineering and LLMs to automate reward design, DRL training, and iterative feedback optimization for humanoid robotics locomotion. STRIDE eliminates the need for task-specific prompt engineering and manual intervention by dynamically integrating environment code, task descriptions, and refinement feedback, achieving significant performance improvements over state-of-the-art methods such as EUREKA and human-designed rewards. Through extensive evaluations on diverse terrains and humanoid tasks, STRIDE demonstrates its ability to generate innovative reward functions, enabling humanoid robots to achieve sprint-level locomotion across complex terrains. Furthermore, STRIDE incorporates human feedback to refine reward alignment without requiring gradient-based optimization, highlighting its potential as a collaborative tool for reward engineering. These results establish STRIDE as a scalable and generalizable solution for automated reward design, advancing the fields of reinforcement learning and humanoid robotics.

\nocite{langley00}

\bibliography{arXiv_stride_tr5}

\begin{thebibliography}{30}
\providecommand{\natexlab}[1]{#1}
\providecommand{\url}[1]{\texttt{#1}}
\expandafter\ifx\csname urlstyle\endcsname\relax
  \providecommand{\doi}[1]{doi: #1}\else
  \providecommand{\doi}{doi: \begingroup \urlstyle{rm}\Url}\fi

\bibitem[Albilani \& Bouzeghoub(2022)Albilani and Bouzeghoub]{albilani2022dynamic}
Albilani, M. and Bouzeghoub, A.
\newblock Dynamic adjustment of reward function for proximal policy optimization with imitation learning: Application to automated parking systems.
\newblock In \emph{2022 IEEE Intelligent Vehicles Symposium (IV)}, pp.\  1400--1408. IEEE, 2022.

\bibitem[Baek et~al.(2024)Baek, Park, Noh, Bae, and Kim]{beak2024chatpcg}
Baek, I.-C., Park, T.-H., Noh, J.-H., Bae, C.-M., and Kim, K.-J.
\newblock Chatpcg: Large language model-driven reward design for procedural content generation.
\newblock In \emph{2024 IEEE Conference on Games (CoG)}, pp.\  1--4, 2024.

\bibitem[Cao et~al.(2024)Cao, Zhao, Cheng, Shu, Chen, Liu, Liang, Zhao, Yan, and Li]{cao2024survey}
Cao, Y., Zhao, H., Cheng, Y., Shu, T., Chen, Y., Liu, G., Liang, G., Zhao, J., Yan, J., and Li, Y.
\newblock Survey on large language model-enhanced reinforcement learning: Concept, taxonomy, and methods.
\newblock \emph{IEEE Transactions on Neural Networks and Learning Systems}, 2024.

\bibitem[Chen et~al.(2024)Chen, Liu, Huang, Wu, Liu, Jiang, Pu, Lei, Chen, Wang, et~al.]{chen2024large}
Chen, J., Liu, Z., Huang, X., Wu, C., Liu, Q., Jiang, G., Pu, Y., Lei, Y., Chen, X., Wang, X., et~al.
\newblock When large language models meet personalization: Perspectives of challenges and opportunities.
\newblock \emph{World Wide Web}, 27\penalty0 (4):\penalty0 42, 2024.

\bibitem[Figueroa et~al.(2024)Figueroa, Tafur, and Kheddar]{figueroa2024reinforcement}
Figueroa, N., Tafur, J., and Kheddar, A.
\newblock Reinforcement learning-based parameter optimization for whole-body admittance control with is-mpc.
\newblock In \emph{2024 IEEE/SICE International Symposium on System Integration (SII)}, pp.\  1405--1410, 2024.

\bibitem[Han et~al.(2024)Han, Yang, Chen, Cai, Chu, and Zhu]{han2024autoreward}
Han, X., Yang, Q., Chen, X., Cai, Z., Chu, X., and Zhu, M.
\newblock Autoreward: Closed-loop reward design with large language models for autonomous driving.
\newblock \emph{IEEE Transactions on Intelligent Vehicles}, 2024.

\bibitem[Jiang et~al.(2024)Jiang, Wang, Shen, Kim, and Kim]{jiang2024survey}
Jiang, J., Wang, F., Shen, J., Kim, S., and Kim, S.
\newblock A survey on large language models for code generation.
\newblock \emph{arXiv preprint arXiv:2406.00515}, 2024.

\bibitem[Katara et~al.(2024)Katara, Xian, and Fragkiadaki]{katara2024gen2sim}
Katara, P., Xian, Z., and Fragkiadaki, K.
\newblock Gen2sim: Scaling up robot learning in simulation with generative models.
\newblock In \emph{2024 IEEE International Conference on Robotics and Automation (ICRA)}, pp.\  6672--6679, 2024.

\bibitem[Kepel \& Valogianni(2024)Kepel and Valogianni]{kepel2024autonomous}
Kepel, D. and Valogianni, K.
\newblock Autonomous prompt engineering in large language models.
\newblock \emph{arXiv preprint arXiv:2407.11000}, 2024.

\bibitem[Kim et~al.(2023)Kim, Baldi, and McAleer]{kim2023language}
Kim, G., Baldi, P., and McAleer, S.
\newblock Language models can solve computer tasks.
\newblock \emph{Advances in Neural Information Processing Systems}, 36:\penalty0 39648--39677, 2023.

\bibitem[Langley(2000)]{langley00}
Langley, P.
\newblock Crafting papers on machine learning.
\newblock In Langley, P. (ed.), \emph{Proceedings of the 17th International Conference on Machine Learning (ICML 2000)}, pp.\  1207--1216, Stanford, CA, 2000. Morgan Kaufmann.

\bibitem[Li et~al.(2024)Li, Yang, Wang, Zhu, Zhou, Qiao, Wang, Li, Lu, and Dai]{li2024auto}
Li, H., Yang, X., Wang, Z., Zhu, X., Zhou, J., Qiao, Y., Wang, X., Li, H., Lu, L., and Dai, J.
\newblock Auto mc-reward: Automated dense reward design with large language models for minecraft.
\newblock In \emph{2024 IEEE/CVF Conference on Computer Vision and Pattern Recognition (CVPR)}, pp.\  16426--16435, 2024.

\bibitem[Li et~al.(2021)Li, Cheng, Peng, Abbeel, Levine, Berseth, and Sreenath]{li2021reinforcement}
Li, Z., Cheng, X., Peng, X.~B., Abbeel, P., Levine, S., Berseth, G., and Sreenath, K.
\newblock Reinforcement learning for robust parameterized locomotion control of bipedal robots.
\newblock In \emph{2021 IEEE International Conference on Robotics and Automation (ICRA)}, pp.\  2811--2817. IEEE, 2021.

\bibitem[Ma et~al.(2024)Ma, Liang, Wang, Huang, Bastani, Jayaraman, Zhu, Fan, and Anandkumar]{ma2024eureka}
Ma, Y.~J., Liang, W., Wang, G., Huang, D.-A., Bastani, O., Jayaraman, D., Zhu, Y., Fan, L., and Anandkumar, A.
\newblock Eureka: Human-level reward design via coding large language models.
\newblock Hybrid, Vienna, Austria, 2024.

\bibitem[Makoviychuk et~al.(2021)Makoviychuk, Wawrzyniak, Guo, Lu, Storey, Macklin, Hoeller, Rudin, Allshire, Handa, et~al.]{makoviychuk2021isaac}
Makoviychuk, V., Wawrzyniak, L., Guo, Y., Lu, M., Storey, K., Macklin, M., Hoeller, D., Rudin, N., Allshire, A., Handa, A., et~al.
\newblock Isaac gym: High performance gpu-based physics simulation for robot learning.
\newblock \emph{arXiv preprint arXiv:2108.10470}, 2021.

\bibitem[Mo et~al.(2022)Mo, Ho, and King]{mo2022managing}
Mo, Y.-W., Ho, C., and King, C.-T.
\newblock Managing shaping complexity in reinforcement learning with state machines - using robotic tasks with unspecified repetition as an example.
\newblock In \emph{2022 IEEE International Conference on Mechatronics and Automation (ICMA)}, pp.\  544--550, 2022.

\bibitem[Nandula et~al.(2021)Nandula, Raj, Deb, and Kumar]{nandula2021neurodynamic}
Nandula, A. K.~R., Raj, S., Deb, A., and Kumar, C.
\newblock A neurodynamic approach to stabilization of a 10 dof biped mechanism using reinforcement learning.
\newblock In \emph{Mechanism and Machine Science: Select Proceedings of Asian MMS 2018}, pp.\  483--496. Springer, 2021.

\bibitem[Naveed et~al.(2023)Naveed, Khan, Qiu, Saqib, Anwar, Usman, Akhtar, Barnes, and Mian]{naveed2023comprehensive}
Naveed, H., Khan, A.~U., Qiu, S., Saqib, M., Anwar, S., Usman, M., Akhtar, N., Barnes, N., and Mian, A.
\newblock A comprehensive overview of large language models.
\newblock \emph{arXiv preprint arXiv:2307.06435}, 2023.

\bibitem[Pattnaik \& Lee(2021)Pattnaik and Lee]{pattnaik2021multitask}
Pattnaik, U. and Lee, M.
\newblock Multi-task transfer with practice.
\newblock In \emph{2021 IEEE Symposium Series on Computational Intelligence (SSCI)}, pp.\  1--8, 2021.

\bibitem[Pei et~al.(2024)Pei, Ma, Zhang, and Li]{pei2024gait}
Pei, R., Ma, P., Zhang, T., and Li, L.
\newblock A gait control method for bipedal robots based on reward function optimization.
\newblock In \emph{2024 4th International Conference on Intelligent Communications and Computing (ICICC)}, pp.\  176--181, 2024.

\bibitem[Qin et~al.(2023)Qin, Chen, Wei, Huang, and Che]{qin2023cross}
Qin, L., Chen, Q., Wei, F., Huang, S., and Che, W.
\newblock Cross-lingual prompting: Improving zero-shot chain-of-thought reasoning across languages.
\newblock \emph{arXiv preprint arXiv:2310.14799}, 2023.

\bibitem[Radosavovic et~al.(2024)Radosavovic, Xiao, Zhang, Darrell, Malik, and Sreenath]{radosavovic2024real}
Radosavovic, I., Xiao, T., Zhang, B., Darrell, T., Malik, J., and Sreenath, K.
\newblock Real-world humanoid locomotion with reinforcement learning.
\newblock \emph{Science Robotics}, 9\penalty0 (89):\penalty0 eadi9579, 2024.

\bibitem[Sarukkai et~al.(2024)Sarukkai, Shacklett, Majercik, Bhatia, R{\'e}, and Fatahalian]{sarukkai2024automated}
Sarukkai, V., Shacklett, B., Majercik, Z., Bhatia, K., R{\'e}, C., and Fatahalian, K.
\newblock Automated rewards via llm-generated progress functions.
\newblock \emph{arXiv preprint arXiv:2410.09187}, 2024.

\bibitem[Self et~al.(2022)Self, Abudia, Mahmud, and Kamalapurkar]{self2022model}
Self, R., Abudia, M., Mahmud, S.~N., and Kamalapurkar, R.
\newblock Model-based inverse reinforcement learning for deterministic systems.
\newblock \emph{Automatica}, 140:\penalty0 110242, 2022.

\bibitem[Singh et~al.(2009)Singh, Lewis, and Barto]{singh2009rewards}
Singh, S., Lewis, R.~L., and Barto, A.~G.
\newblock Where do rewards come from.
\newblock In \emph{Proceedings of the annual conference of the cognitive science society}, pp.\  2601--2606. Cognitive Science Society, 2009.

\bibitem[Tan \& Wang(2024)Tan and Wang]{tan2024dynamic}
Tan, J. and Wang, Y.
\newblock Dynamic inverse reinforcement learning for feedback-driven reward estimation in brain machine interface tasks.
\newblock In \emph{2024 46th Annual International Conference of the IEEE Engineering in Medicine and Biology Society (EMBC)}, pp.\  1--4, 2024.

\bibitem[Tang et~al.(2024)Tang, Hiraoka, Hiraoka, Shi, Kawaharazuka, Kojima, Okada, and Inaba]{tang2024humanmimic}
Tang, A., Hiraoka, T., Hiraoka, N., Shi, F., Kawaharazuka, K., Kojima, K., Okada, K., and Inaba, M.
\newblock Humanmimic: Learning natural locomotion and transitions for humanoid robot via wasserstein adversarial imitation.
\newblock In \emph{2024 IEEE International Conference on Robotics and Automation (ICRA)}, pp.\  13107--13114, 2024.

\bibitem[Weng et~al.(2021)Weng, Hashemi, and Arami]{weng2021natural}
Weng, J., Hashemi, E., and Arami, A.
\newblock Natural walking with musculoskeletal models using deep reinforcement learning.
\newblock \emph{IEEE Robotics and Automation Letters}, 6\penalty0 (2):\penalty0 4156--4162, 2021.

\bibitem[Zare et~al.(2024)Zare, Kebria, Khosravi, and Nahavandi]{zare2024survey}
Zare, M., Kebria, P.~M., Khosravi, A., and Nahavandi, S.
\newblock A survey of imitation learning: Algorithms, recent developments, and challenges.
\newblock \emph{IEEE Transactions on Cybernetics}, 2024.

\bibitem[Zeng et~al.(2024)Zeng, Mu, and Shao]{zeng2024learning}
Zeng, Y., Mu, Y., and Shao, L.
\newblock Learning reward for robot skills using large language models via self-alignment.
\newblock \emph{arXiv preprint arXiv:2405.07162}, 2024.

\end{thebibliography}
\bibliographystyle{icml2025}

\newpage
\appendix
\onecolumn
\section{FULL PROMPTS}
In this section, we provide all STRIDE prompts. At a high level, STRIDE combines agentic engineering principles and large language models to automate reward design, RL training, and feedback optimization for humanoid robotics locomotion. Below, we present the general system prompt and feedback-based optimization prompt used in STRIDE.

\subsection*{Prompt 1: Initial System Prompt}
\lstset{ 
	basicstyle=\ttfamily\small, 
	breaklines=true, 
	frame=single, 
	backgroundcolor=\color{lightgray}, 
	linewidth=\textwidth 
}
\begin{lstlisting}
	You are a highly skilled reward engineer tasked with designing effective, efficient, and robust reward functions for reinforcement learning tasks. Your goal is to create a reward function that guides the agent to learn the task described in text. To achieve this, adhere to the following guidelines:
	
	1. Dynamic Use of Observations:
	a) The reward function must dynamically adapt to the current environment's observation space. Utilize all available observation variables provided by the environment code without introducing undefined variables.
	b) Ensure the function can accommodate changes in the observation space without manual modifications.
	
	2. Task-Oriented Reward Design:
	a) Clearly focus on optimizing the task described in {task_description}. For example, maximize forward velocity while minimizing energy consumption and maintaining stability.
	b) Use a multi-objective framework where appropriate, balancing competing goals using weighted components.
	
	3. Code Robustness and Compatibility:
	a) Ensure the reward function is TorchScript-compatible (e.g., using `torch.Tensor` and avoiding operations incompatible with TorchScript).
	b) Validate input dimensions, types, and device consistency to prevent runtime errors.
	
	4. Efficiency and Resource Optimization:
	a) Avoid redundant computations and minimize resource usage. Use vectorized tensor operations whenever possible for better performance.
	b) Introduce necessary transformations (e.g., normalization or scaling) to keep reward values within a manageable range.
	
	5. Encourage Innovation:
	a) Explore advanced reward strategies, such as:
	b) Dynamic adjustment of reward weights based on the agent's performance.
	c) Incorporating contrastive reward signals (e.g., rewarding the agent for behavior that contrasts with failures).
	d) Adaptive rewards based on the agent's learning stage.
	
	The reward function signature can be: {task_reward_signature_string}.
	
	Since the reward function will be decorated with `@torch.jit.script`, ensure all tensors and variables introduced are compatible with TorchScript and reside on the same device as the input tensors.
	
	Example prompt structure:
	```python
	@torch.jit.script
	def compute_reward(
	kwargs
	) -> Tuple[torch.Tensor, Dict[str, torch.Tensor]]:
	# Reward computation logic here, dynamically utilizing the provided observation data
	
\end{lstlisting}

\subsection*{Prompt 2: Initial User Prompt}
\begin{lstlisting}
	The Python environment is {task_obs_code_string}. Write a reward function for the following task: {task_description}.
	
	Your reward function must adhere to the following guidelines to ensure high code success rates and minimal errors:
	
	1. Dynamic Use of Observations:
	1.1 The reward function must dynamically adapt to the current environment's observation space. Utilize all available observation variables provided by the environment code without introducing undefined variables.  
	1.2 Ensure the function can accommodate changes in the observation space without manual modifications.
	
	2. Task-Oriented Reward Design for Bolt-like Running:
	2.1 Focus on optimizing the humanoid robot's running posture and speed to mimic Usain Bolt's sprinting characteristics.  
	2.2 Use the humanoid's URDF parameters (e.g., joint angles, joint order, masses, dimensions) and observation data to design the reward function.  
	2.3 Key aspects of Bolt's running to replicate:
	2.3.1 Lower Body Trajectory:
	2.3.1.1 Hip Joint: Forward swing (60-80 degrees), backward extension (-10 to -20 degrees).  
	2.3.1.2 Knee Joint: High flexion during swing (90-120 degrees), slight flexion during stance (20-40 degrees).  
	2.3.1.3 Ankle Joint: Toe-off with plantar flexion (10-15 degrees), slight dorsiflexion at landing (0-5 degrees).  
	2.3.2 Upper Body Coordination:
	2.3.2.1 Shoulders: Arm swing synchronized with leg movements (45-60 degrees).  
	2.3.2.2 Elbows: Maintain 90-120 degrees flexion for balance and propulsion.  
	2.3.3 Torso Position:
	2.3.3.1 Maintain a forward tilt of 20-25 degrees for stability and momentum.  
	2.3.4 Speed and Step Dynamics:
	2.3.4.1 Achieve step lengths of 2.5-3 meters with optimized step frequency.  
	2.4 Integrate penalties for deviations from the target trajectory, unnecessary energy consumption, or instability.
	
	3. URDF Joint Order and Initialization:
	3.1 The reward function must respect the joint order defined in the URDF file. Dynamically extract and utilize the joint names and their corresponding indices.  
	3.2 Ensure the robot's initial pose aligns with the URDF-defined default (e.g., standing position) and adjust the joint positions for starting actions (e.g., running stance).  
	3.3 Validate joint order consistency when applying target angles or computing rewards:  
	```python  
	target_angles = {  
		"hip_joint": 0.5,  
		"knee_joint": 1.0,  
		"ankle_joint": -0.3  
	}  
	for joint_name, target in target_angles.items():  
	joint_index = joint_names.index(joint_name)  
	current_angle = p.getJointState(robot_id, joint_index)[0]  
	reward -= abs(current_angle - target)  # Penalize deviations  
	```
	
	4. Transformation and Normalization:
	4.1 Normalize each reward component to a fixed range (e.g., [0, 1]) using transformations like `torch.exp` or `torch.tanh`.  
	4.2 Introduce temperature parameters to control the scaling of transformed components. These parameters must be defined inside the function.
	
	5. Code Efficiency and Robustness:
	5.1 Optimize computations by avoiding redundant operations. Use vectorized tensor calculations wherever possible.  
	5.2 Ensure compatibility with TorchScript:  
	5.2.1 Use `torch.Tensor` for all variables.  
	5.2.2 Validate that all tensors are on the same device (CPU/GPU).
	
	6. Debugging and Interpretability:
	6.1 Provide an optional debugging mode that logs intermediate values of reward components and highlights potential issues during execution.  
	6.2 Use meaningful keys in the reward dictionary for easy identification of each component.
	
	7. Incorporate Advanced Strategies:
	7.1 Dynamic reward weighting: Adjust the importance of components like speed, stability, and energy efficiency based on training progress.  
	7.2 Contrastive reward signals: Reward behavior that contrasts with failure scenarios to accelerate learning.  
	7.3 Multi-objective optimization: Balance speed, efficiency, and stability in a cohesive reward framework.
	
	### Additional Requirements for Running Pose:
	8.1 The reward function should ensure that the humanoid robot's arm and leg movements are synchronized in a natural way, mimicking the front-and-back swinging of limbs as seen in Usain Bolt's running style.  
	8.2 Ensure the robot's legs do not swing inward (no "knock knees"), and the arms and legs should move in opposite directions, maintaining a realistic sprinting posture.  
	8.3 Integrate a penalty for improper arm-leg coordination or unnatural limb movements (e.g., arm and leg moving in the same direction).
	
	### Example Reward Function Format:
	```python
	@torch.jit.script
	def compute_reward(
	kwargs
	) -> Tuple[torch.Tensor, Dict[str, torch.Tensor]]:
	# Reward computation logic
	total_reward = ...
	reward_components = {
		"speed": ...,  # Speed optimization based on Bolt's step length and frequency
		"stability": ...,  # Penalize deviations from balanced posture
		"energy_efficiency": ...,  # Encourage minimal energy consumption
		"trajectory_alignment": ...,  # Match joint angles to Bolt-like trajectories
		"arm_leg_coordination": ...,  # Reward for proper arm-leg synchronization
	}
	return total_reward, reward_components
	```
	
\end{lstlisting}

\subsection*{Prompt 3: Reward Reflection and Feedback}
\begin{lstlisting}
	We trained a reinforcement learning (RL) policy using the provided reward function code and tracked the following data:
	a) Values of the individual reward components (maximum, mean, minimum) at every {epoch_freq} epochs.
	b) Global policy metrics, such as success rates and episode lengths.
	
	### Guidelines for Analysis and Improvement:
	1. Analyze Tracked Data:
	a) Compare reward component trends across epochs:
	b) Identify components with low variance or near-constant values, as they may not provide useful gradients.
	c) Examine components with extreme values (e.g., very large maximums) that may dominate the overall reward.
	d) Correlate reward component trends with global policy metrics:
	e) Determine whether increases in certain reward components correlate with improved success rates or reduced episode lengths.
	f) Check for stagnation:
	g) If success rates plateau or remain near zero, identify potential bottlenecks in the reward design.
	
	2. Optimize and Debug:
	a) For ineffective reward components (e.g., low variance or uncorrelated with policy improvement), consider:
	b) Adjusting their scale or introducing normalization.
	c) Rewriting or removing components that contribute little to policy improvement.
	d) For dominant components, rescale them to balance their influence relative to other components.
	
	3. Innovate with Advanced Strategies:
	a) Introduce dynamic reward weighting:
	b) Adjust the importance of each reward component based on the agent's performance and learning stage.
	c) Explore multi-objective optimization:
	d) Combine competing goals (e.g., speed vs. energy efficiency) using weighted components.
	e) Apply contrastive analysis:
	f) Compare successful and failed episodes to identify and reward key differences.
	
	4. Leverage Visualization and Tools:
	a) Use visualization tools (e.g., TensorBoard, matplotlib) to plot:
	b) Reward components over time.
	c) Success rates and episode lengths across epochs.
	d) Highlight trends or anomalies to guide further optimization.
	
	5. Iterate and Improve:
	a) After analyzing the data, rewrite the reward function with improvements:
	b) Balance competing objectives.
	c) Address stagnation or inefficiencies.
	d) Ensure that the revised function is efficient, robust, and aligned with the training goals.
	
	### Example Improvement Workflow:
	1. Analyze tracked data and identify issues (e.g., unbalanced components, stagnant success rates).
	2. Propose specific adjustments to the reward function (e.g., rescaling, adding/removing components).
	3. Rewrite the reward function and validate it through further training.
	4. Use visualization and tools to evaluate the effectiveness of changes.
	
	Your goal is to iteratively refine the reward function to maximize success rates, reduce episode lengths, and ensure robust policy learning. Please provide the updated reward function code in the following format:
	```python
	@torch.jit.script
	def compute_reward(
	kwargs
	) -> Tuple[torch.Tensor, Dict[str, torch.Tensor]]:
	# Improved reward computation logic
	total_reward = ...
	reward_components = {
		"speed": ...,
		"stability": ...,
		"energy_efficiency": ...,
	}
	return total_reward, reward_components
	
	Please carefully analyze the policy feedback and provide a new, improved reward function that can better solve the task. Follow these steps to analyze and improve the reward function:
	
	1. Policy Feedback Analysis:
	a) Analyze the success rates and reward trends over time:
	b) If the success rates remain near zero, identify which reward components may not provide useful gradients and rewrite the function entirely.
	c) If certain reward components exhibit near-identical values (low variance), diagnose their inefficiency. This may involve:
	d) Adjusting their scaling or temperature parameters.
	e) Redesigning the reward component for better optimization.
	f) Discarding components that contribute negligible gradients.
	g) If some reward components dominate in magnitude, normalize them to an appropriate range to ensure balanced optimization.
	
	2. Advanced Reward Design:
	a) Incorporate innovative reward strategies:
	b) Use contrastive reward signals by comparing successful and failed behaviors.
	c) Implement dynamic weighting: Adjust the importance of reward components based on the agent's training progress.
	d) Consider multi-objective optimization frameworks for tasks with competing goals (e.g., speed vs. energy efficiency).
	
	3. Structured Debugging:
	a) Use diagnostic tools to analyze the reward function's effectiveness:
	b) Plot reward trends and success rates over training epochs.
	c) Log and visualize the contribution of each reward component to the total reward.
	d) Identify bottlenecks or unexpected behavior in the agent's policy by correlating rewards with actions.
	
	4. Efficiency and Robustness:
	a) Avoid redundant computations and unnecessary variable creation.
	b) Ensure compatibility with TorchScript, adhering to the environment's input-output constraints.
	c) Validate input types, shapes, and device consistency to prevent runtime errors.
	
	### Example Reward Function Signature:
	```python
	@torch.jit.script
	def compute_reward(
	kwargs
	) -> Tuple[torch.Tensor, Dict[str, torch.Tensor]]:
	# Reward computation logic here, dynamically utilizing feedback analysis
	
	
\end{lstlisting}

\subsection*{Prompt 4: Code formatting tip}
\begin{lstlisting}
	The output of the reward function should consist of two items:
	(1) The total reward (a scalar value representing the overall task performance).
	(2) A dictionary of individual reward components, where each key represents a specific component's name and value is its contribution.
	
	The code output should be formatted as a Python code string: "```python ... ```".
	
	#### Helpful Tips for Writing the Reward Function:
	1. Reward Component Design:
	a) Decompose the reward into meaningful components that represent specific task objectives (e.g., speed, stability, energy efficiency).
	b) Balance competing objectives using appropriate weighting or normalization techniques.
	c) Include a brief comment for each reward component to explain its purpose and contribution.
	
	2. Dynamic Adaptation:
	a) Use only the observation variables provided by the environment (`self.` attributes) to design the reward function. Do not introduce new input variables.
	b) Dynamically adapt to the available observation space, ensuring compatibility with changes in the environment.
	
	3. Transformation and Normalization:
	a) Normalize each reward component to a fixed range (e.g., [0, 1]) using transformations like `torch.exp` or `torch.tanh`.
	b) If you apply transformations, introduce a temperature parameter for each component. Ensure this parameter is defined inside the function and not passed as an input variable.
	
	4. Code Efficiency and Robustness:
	a) Optimize computations by avoiding redundant operations. Use vectorized tensor calculations wherever possible.
	b) Ensure compatibility with TorchScript:
	c) Use `torch.Tensor` for all variables.
	d) Validate that all tensors are on the same device (CPU/GPU).
	
	5. Debugging and Interpretability:
	a) Provide an optional debugging mode that outputs each reward component's value for analysis.
	b) Use meaningful keys in the reward dictionary for easy identification of each component.
	
	6. Encourage Innovation:
	a) Consider advanced reward design strategies:
	b) Dynamic reward weighting: Adjust the importance of each reward component based on the agent's performance or learning stage.
	c) Contrastive reward signals: Reward behavior that contrasts with failed attempts to accelerate learning.
	d) Multi-objective optimization: Combine multiple objectives in a way that promotes balanced learning.
	
	#### Example Reward Function Format:
	```python
	@torch.jit.script
	def compute_reward(
	kwargs
	) -> Tuple[torch.Tensor, Dict[str, torch.Tensor]]:
	# Reward computation logic here
	total_reward = ...
	reward_components = {
		"speed": ...,
		"stability": ...,
		"energy_efficiency": ...,
	}
	return total_reward, reward_components
	
	
\end{lstlisting}


\end{document}